\documentclass[accepted]{uai2021}

\usepackage{natbib} 
\usepackage{mathtools} 

\usepackage[utf8]{inputenc} 
\usepackage{hyperref}       
\usepackage{url}            
\usepackage{booktabs}       
\usepackage{amsfonts}       
\usepackage{nicefrac}       
\usepackage{microtype}      
\usepackage{easyeqn}
\usepackage{bbm}
\usepackage[ruled,vlined]{algorithm2e}
\usepackage{xcolor}
\usepackage{graphicx}
\usepackage{subcaption}

\usepackage{xr}
\DeclareMathAlphabet{\mathcal}{OMS}{cmsy}{m}{n}
\DeclareMathOperator{\E}{\mathop{\mathbb{E}}}

\newcommand\R{\mathbb{R}}
\renewcommand\O{\mathcal{O}}

\renewcommand\vec{\boldsymbol}
\newcommand\set{\mathcal}

\usepackage{pifont}
\newcommand{\cmark}{\color{green}{\ding{51}}}%
\newcommand{\xmark}{\color{red}{\ding{55}}}%

\title{Tensor-Train Density Estimation}

\author[1]{\href{mailto:Georgii Novikov <georgii.novikov@skoltech.ru>?Subject=Tensor-Train Density Estimation}{Georgii~S.~Novikov}{}} 
\author[1]{Maxim~E.~Panov}
\author[1]{Ivan~V.~Oseledets}

\affil[1]{%
  Skolkovo Institute of Science and Technology \\
  Moscow, Russia
}

\begin{document}
  \maketitle

  \begin{abstract}
    Estimation of probability density function from samples is one of the central problems in statistics and machine learning. Modern neural network-based models can learn high dimensional distributions but have problems with hyperparameter selection and are often prone to instabilities during training and inference. We propose a new efficient tensor train-based model for density estimation (TTDE). Such density parametrization allows exact sampling, calculation of cumulative and marginal density functions, and partition function. It also has very intuitive hyperparameters. We develop an efficient non-adversarial training procedure for TTDE based on the Riemannian optimization. Experimental results demonstrate the competitive performance of the proposed method in density estimation and sampling tasks, while TTDE significantly outperforms competitors in training speed.
  \end{abstract}

  \section{Introduction}
  \label{sec:introduction}
  In this paper, we consider a problem of nonparametric density estimation, which is one of the central problems in statistics and machine learning. Recent progress in the development of artificial neural networks has given rise to many new methods of solving this problem, including variational autoencoders (VAE, \cite{Kingma2013}), generative adversarial networks (GAN, \cite{goodfellow2014generative}), autoregressive neural networks~\citep{Oord2016}, invertible flows~\citep{Dinh2016} among some others. These methods allow us to overcome the curse of dimensionality and make it possible to estimate the density of such high-dimensional and nontrivial data as images and sound. However, all these new approaches lack the simplicity and interpretability of the classical kernel density estimation method~\citep{Scott1977}. On the other hand, kernel density estimation usually performs poorly even in moderate dimensions~\citep{Wang2019}.

  This paper aims to build a new method of nonparametric density estimation: tensor-train density estimation (TTDE).
  The idea is to construct a tensor-train approximation to the coefficients' matrix for the expansion of the density function in some basis.
  We will show that an approximation in this parametric form has several important features that other models do not have (at least not simultaneously): exact sampling, ability to calculate cumulative density function and exact calculation of partition function. Moreover, we propose an efficient training procedure based on Riemannian optimization, which is easy to implement and avoids the problems of instability typical for the methods based on adversarial training.

  \paragraph{Contributions of this work.}
  Although extremely powerful and effective, modern neural network-based models have their drawbacks. Some of them do not have tractable log-likelihood at all (like GANs) or have only surrogates for it (lower bound for VAEs, unnormalized log-density for energy-based models). Other methods can not sample from the trained distribution or require the whole additional sampling procedure like MCMC and thus can generate only approximate samples (energy-based models, BNAF~\citep{DeCao2019}). Many powerful models require ``middle-men'' during the training process (discriminator for adversarial models, MCMC sampling for energy-based models), which significantly complicates the development and analysis of such models. These properties for different methods are summarized in Table~\ref{tab:comp}.

  \begin{table*}
    \centering
    \caption{Comparison of the capabilities of different density estimation models. *FFJORD does not use true log-likelihood in the training process and instead uses its unbiased estimate.} \label{tab:comp}
    \begin{tabular}{ccccc}
      \toprule 
      \bfseries Method & \bfseries Exact Sampling & \bfseries Tractable LL & \bfseries No middle-man Training & \bfseries Computation of CDF \\
      \midrule 
      FFJORD & \cmark & \cmark* & \cmark* & \xmark\\
      Normalizing Flows & \cmark & \cmark & \cmark  & \xmark\\
      GANs & \cmark & \xmark & \xmark  & \xmark\\
      VAEs & \cmark & \xmark & \cmark  & \xmark\\
      Autoregressive & \cmark & \cmark & \cmark  & \xmark\\
      Energy-based & \xmark & \xmark & \xmark  & \xmark\\
      \midrule
      TTDE (ours) & \cmark & \cmark & \cmark & \cmark \\
      \bottomrule 
    \end{tabular}
    \
  \end{table*}

  Neural network-based methods are famous for their strong dependence on the choice of hyperparameters, architecture and optimization method. We believe that there is a gap between simple, intuitive models for low-dimensional data and powerful, capable of solving most difficult tasks, yet very fragile and hard to theoretically analyze neural network-based methods. In this work, we try to fill this gap.

  The main contributions of our work are as follows.
  \begin{itemize}
    \item We propose a new generative tensor-based approach \textit{tensor-train density estimation (TTDE)} that allows fast sampling and efficient computation of functionals of probability density function.

    \item We show that TTDE can be trained using Riemannian optimization targeting a variety of different functionals, including those that are intractable for previously existing models (namely, direct $L_2$ loss between target probability $p(x)$ and approximation $q_{\vec{\theta}}(x)$).

    \item We illustrate the competitive performance of our approach on a series of examples.
  \end{itemize}

  \section{Why tensor-train is good for density approximation?}
  \label{sec:theory}

  \subsection{Problem statement}
  Suppose we are given i.i.d. samples $\vec{x}^{(1)}, \ldots, \vec{x}^{(N)}, \quad \vec{x}^{(i)} \in \mathbb{R}^d, i=1, \ldots, N$ from an unknown probability distribution with a density $p(\vec{x})$. We want to find an approximation to this density. It is typically done by using some family of functions:
  \begin{EQA}[c]
    \label{eq:general-approx}
    p(\vec{x}) \approx q_{\vec{\theta}}(\vec{x}),
  \end{EQA}
  where \(q_{\vec{\theta}} \in \set{Q} = \{ q_{\vec{\theta}} \}_{\vec{\theta} \in \Theta}, \Theta \subset \R^D\).

  To perform approximation~\eqref{eq:general-approx}, some measure of discrepancy between probability densities $p$ and $q_\theta$ should be computed (given only samples from $p$) and then optimized with respect to $\theta$.

  In this paper, we propose to use densities represented in the \emph{low-rank tensor-train format} as $\set{Q}$. This approach has been shown to be successful in~\citep{Dolgov2020} which targeted the problem of the computationally efficient approximation to the given density. This problem is very different from the problem of density estimation from samples which we consider in our work. We aim to fill this gap by developing a systematic approach for sample-based training of such models.

  The general approach in non-parametric statistics is firstly to choose some basis functions \(\Phi(\vec{x}) = \{f_k(\vec{x})\}_{k = 1}^K\) (e.g B-splines or Fourier series) and then search for the approximation in the linear space induced by this basis:
  \begin{EQA}[c]
    q_{\vec{\theta}}(\vec{x}) = \bigl\langle \alpha_{\vec{\theta}}, \Phi(\vec{x}) \bigr\rangle = \sum_{k = 1}^K \alpha_{\vec{\theta}, k} f_k(\vec{x}),
  \end{EQA}
  where usually the coefficients vector $\alpha_{\vec{\theta}}$ simply coincides with the parameter vector $\vec{\theta}$, i.e. $\alpha_{\vec{\theta}} \equiv \vec{\theta}$.

  One of the standard ways to build a multidimensional basis is to take a Cartesian product of several one-dimensional bases, i.e., setting \(f_{i_1, \cdots, i_d}(\vec{x}) = f_{i_1}(x_1) \cdots f_{i_d}(x_d)\) for \(d\)-dimensional input \(\vec{x}\) and some functions \(f_{1}, \dots, f_{d}\colon \R \to \R\). In this case, the coefficients vector \(\alpha_{\vec{\theta}}\) becomes structured as a \(d\)-dimensional tensor $\alpha_\theta \in \R^{K^d}$ whose size grows exponentially with the dimension. In this work, we propose to consider only a low-rank subspace of the linear span, i.e., functions $q_{\theta}(x)$, weight tensor $\alpha_\theta$ of which can be represented in low-rank tensor-train format. Such a representation allows achieving linear in the dimension computational and storage costs for operations such as calculation of a function at a point, differentiation and integration.

  \subsection{Proposed representation of the density}
  \label{sec:basis}

  \paragraph{Tensor-product basis.}
  Consider a basis set of 1-dimensional functions $\set{B} = \{f_i\}_{i=1}^{m}, f_i\colon \R \rightarrow \R$. The construction of the $d$-dimensional basis set can be done on top of $\set{B}$ as follows:
  \begin{EQA}[c]
    \set{B}^{(d)} = \bigl\{ f_{i_1, \dots, i_d} \bigr\}_{i_1=1, \dots, i_d=1}^{m, \dots, m},
  \end{EQA}
  where
  \begin{EQA}[c]
    f_{i_1, \dots, i_d}(\vec{x}) = f_{i_1}(x_1) \cdots f_{i_d}(x_d)\colon \R^{d} \rightarrow \R.
  \end{EQA}
  We are going to approximate the target distribution via the linear function expansion in this basis:
  \begin{EQA}[rcl] \label{eq:approx-of-p}
  q_{\vec{\theta}}(x) \in \set{Q} & = & \text{span} ~ \set{B}^{(d)},
  \\
  q_\theta(\vec{x}) & = & \sum_{i_1=1, \dots, i_d=1}^{m, \dots, m} \alpha_{\vec{\theta}}^{i_1, \dots, i_d} f_{i_1, \dots, i_d}(\vec{x)} \\
  & = & \sum_{i_1=1, \dots, i_d=1}^{m, \dots, m} \alpha_{\vec{\theta}}^{i_1, \dots, i_d} f_{i_1}(x_1) \cdots f_{i_d}(x_d) \\
  & = & \sum_{i_1=1, \dots, i_d=1}^{m, \dots, m} \alpha_{\vec{\theta}}^{i_1, \dots, i_d} \Phi_{i_1, \dots, i_d}(\vec{x}) = \langle \alpha_{\vec{\theta}}, \Phi(\vec{x}) \rangle,
  \end{EQA}
  where the tensor $\Phi(\vec{x})$ is a \emph{rank-1 feature map} defined by
  \begin{EQA}[rcl]
    \Phi(\vec{x}) & = & \vec{f}(x_1) \otimes \cdots \otimes \vec{f}(x_d), \\
    \vec{f}(x) & = & \bigl(f_1(x), \dots, f_m(x)\bigr).
  \end{EQA}
  Such a feature map was previously used for the classification problems with tensor-based models in~\citep{Cohen2015,Khrulkov2017,Stoudenmire}.

  \paragraph{Tensor-train format.}
  In the form above, $\alpha_{\vec{\theta}}$ is a very large tensor of size $m^d$. To be able to store and interact with this tensor in high dimensions, we will work with the tensors $\alpha_{\vec{\theta}}$, which can be represented in the tensor-train (TT) format:
  \begin{EQA}[rcl]
    \label{eq:tt-of-alpha}
    \alpha_{\vec{\theta}}^{i_1 \dots i_d} & = & G_1[\cdot, i_1, \cdot] G_2[\cdot, i_2, \cdot] \cdots G_d[\cdot, i_d, \cdot].
  \end{EQA}
  Here $G_i$ are the so-called cores of the tensor-train decomposition, which are $3$-dimensional tensors of size $\left[r_{i - 1} \times m \times r_{i}\right]$ (with condition $r_0 = r_{d} = 1$). Here $G_i[\cdot, a, \cdot]$ represents a matrix which is the $a$-th slice of the core $G_i$ along the second axis. The vector $\vec{r} = (r_1, \dots, r_{d - 1})$ is called the vector of TT-ranks. Further in the text, unless otherwise stated, we will use a single natural number $r$ to refer to the TT-decomposition with rank $\vec{r} = (r, \dots, r)$.

  \paragraph{Computation in the TT-format.}
  When all the coefficients are known, the inner product of two tensors in the TT-format can be computed efficiently. This procedure is summarized in Algorithm~\ref{alg:inner_product}. It requires $\O(d m r_1^2 r_2)$ operations to compute, where $r_1$ is the maximum of two ranks of given tensors, $r_2$ is the minimum of two ranks, $d$ is the number of dimensions of the tensors, and $m$ is the maximum size of the dimensions. For example, the evaluation of the function  $q_{\vec{\theta}}$ at a point $\vec{x}$ according to~\eqref{eq:approx-of-p} is an inner-product of the weight tensor with a rank-1 tensor and thus requires only $\O(d m r^2)$ time. In a similar way, we can marginalize out some dimensions and therefore compute marginal densities of $q_{\vec{\theta}}$:
  \begin{EQA}
    \label{eq:marginals}
    q_{\vec{\theta}}(x_1, \cdots, x_{k - 1}) = \Bigl\langle \alpha_{\vec{\theta}}, \Phi(x_1, \cdots, x_{k - 1}) \otimes \int \Phi(x_k, \cdots, x_d) \Bigr\rangle,
  \end{EQA}
  where
  \begin{EQA}[rcl]
    \Phi(x_1, \cdots, x_{k - 1}) & = & \vec{f}(x_1) \otimes \cdots \otimes \vec{f}(x_{k - 1}), \\
    \int \Phi(x_{k}, \cdots, x_d) & = & \left( \int \vec{f}(x_{k}) dx_{k} \right) \otimes \cdots \otimes \left( \int \vec{f}(x_d) dx_d \right)
  \end{EQA}
  or even calculate cumulative density function along some dimension $k$:
  \begin{EQA}
    \label{eq:cumulatives}
    && q_{\vec{\theta}}(x_1, \cdots, x_{k - 1}, x_k < A) = \\
    && \Bigl\langle \alpha_{\vec{\theta}}, \Phi(x_1, \cdots, x_{k - 1}) \otimes \int_{-\infty}^{A} f(x_k) dx_k \otimes \int \Phi(x_{k + 1}, \cdots, x_d) \Bigr\rangle.
  \end{EQA}

  \begin{algorithm}[t]
    \SetAlgoLined
    \KwResult{Inner product of tensors $T_1$ and $T_2$ represented in TT-format with cores $\{G^{(1)}_i\}_{i=1}^d$ and $\{G^{(2)}_i\}_{i=1}^d$ and ranks $r_1$ and $r_2$, respectively}
    Initialize $res$ with $[1 \times 1]$ identity matrix \;
    \For{$p \gets 1$ to $d$}{
      $res \gets \text{einsum}('ix,inj,xny \rightarrow jy', res, G^{(1)}_p, G^{(2)}_p)$ \;
    }
    \Return{$res$}\;
    \caption{Multiplication of two tensors represented in tensor-train format. On each step, we store the contraction of the two prefixes of the lists of cores. Each such contraction can be updated from the previous step in $\O\bigl(\max(r_1^2, r_2^2) \min(r_1, r_2)\bigr)$ time, which gives $\O\bigl(d \max(r_1^2, r_2^2) \min(r_1, r_2)\bigr)$ complexity of the full product.}
    \label{alg:inner_product}
  \end{algorithm}

  \paragraph{Squared TTDE.} \label{par:sqr-ttde}
  The main drawback of proposed model is that it's not guaranteed to be always non-negative. If the expressivity of the model is sufficient, we expect the resulting approximation to be sufficiently similar to the real distribution, and hence the negative regions can be neglected. However, it is too optimistic to expect such behavior for complex high-dimensional distributions. If we want to apply the model to medium dimensions, we have to overcome this problem. One solution is to require that all the kernels in the tensor train~\eqref{eq:tt-of-alpha} are non-negative. Then, if all basis functions are also always non-negative (which can be assumed by construction), then the resulting function $q_\theta$ will be non-negative everywhere. Notice, that this approach will require additional changes to the optimization process. We propose another modification of our model -- \emph{squared TTDE}. Instead of approximation~\eqref{eq:tt-of-alpha} we suggest to use squared version of it:
  \begin{EQA}[c]
    q_\theta(\vec{x}) = \Bigl\langle \alpha_\theta, \Phi(\vec{x}) \Bigr\rangle^2,
  \end{EQA}
  which automatically implies non-negativity. In this form, some operations will become more computationally expensive (e.g., calculating marginals), but still manageable, while the model can be trained using the classical method of likelihood maximization. Experiments with the squared TTDE are presented in Section~\ref{subsec:real-world} and summarized in Table~\ref{tab:sqrttde-results}.

  \subsection{Sampling}
  Direct application of the trained density model is sampling from that model. GANs, for example, can not do anything but sampling (and require one forward pass through the network to do it). Models based on normalizing flows can both infer density and sample (and in general require one forward pass as well). Energy-based models can not generate exact samples from their learned densities and moreover, require an additional iterative procedure to get approximate samples (like Markov Chain Monte Carlo sampling).

  Probability density function represented in the tensor-train format allows fast, exact sampling in the autoregressive fashion: as we can calculate marginals of $q_{\vec{\theta}}(x)$, we can take sample seeds $\vec{u} \sim \mathcal{U}([0; 1]^d)$ and then sample coordinates of $\vec{x}$ one by one such that $q_{\vec{\theta}}(\xi_1 < x_1) = u_1$, $q_{\vec{\theta}}(\xi_2 < x_2 \mid \xi_1 = x_1) = u_2$ and so on. On a step $k$ to sample $x_k$, we assume that we already know all $x_i, i < k$ and thus should find such a number $A$, that
  \begin{EQA}[c]
    q_{\vec{\theta}}(\xi_k < A \mid \{\xi_i = x_i\}_{i < k})
    = \frac{q_{\vec{\theta}}(\xi_k < A, \{\xi_i = x_i\}_{i < k})}{q_{\vec{\theta}}(\{\xi_i = x_i\}_{i < k})} = u_k,
  \end{EQA}
  which is a simple $1$-dimensional search on a monotonically increasing function (up to an approximation error) and thus can be performed with any appropriate algorithm (e.g. binary search).

  TT-format allows an efficient implementation of the described algorithm. Notice that the cumulative density function value can be computed in a cycle over dimensions, where each iteration \(k\) of which can be decomposed in four steps:
  \begin{enumerate}
    \item Contraction of $\Phi(x_1, \cdots, x_{k - 1})$ with cores $G_1, \cdots, G_{k - 1}$. Let us call it $Q^{left}_k$, which is a vector of size $r_{k - 1}$.

    \item Contraction of $\int \Phi(x_{k + 1}, \cdots, x_d) d x_{k + 1} \cdots d x_d$ with cores $G_{k + 1}, \cdots, G_{d}$. Lets call it $Q^{right}_k$, which is a vector of size $r_{k}$.

    \item Multiplication of core $G_k$ with $Q^{left}_k$ and $Q^{right}_k$ along it's left and right dimensions, which results in a vector $Q^{inner}_k$ of size $m$, which is the size of the 1-dimensional basis.

    \item Dot product of $Q^{inner}_k$ with the vector $\int_{-inf}^{x_k} f(x_k) dx_k$.
  \end{enumerate}
  In the considered algorithm, $Q^{right}_k$ can be precomputed beforehand. $Q^{left}_k$ depends only on $x_i$ for $i < k$, and consequently, $Q^{left}_{k + 1}$ can be derived from $Q^{left}_k$ by one matrix-vector product. Thus, steps 1, 2 and 3 require only $\O(d m r^2)$ operations in total. That means that one calculation of the cumulative density function during the $1$-dimensional search boils down to a cheap calculation of $\int_{-inf}^{x_k} \vec{f}(x_k) dx_k$ and one vector-vector product of size $m$. In total, this algorithm requires $O(d m r^2 + d m L)$ operations, where $L$ is a number of iterations that a  $1$-dimensional search would do (for binary search $L$ would be proportional to the log of the desired result precision). The resulting sampling procedure is summarized in Algorithm~\ref{alg:sampling}.

  \begin{algorithm}[t]
    \KwResult{Sample $\vec{x} \sim q_{\vec{\theta}}$}

    Sample $\vec{u} \sim U([0; 1]^d)$ \;
    Precompute $Q^{right}_k$ for all $k$ \;
    Initialize $Q^{left}_1$ with $1$ \;

    \For{$k \gets 1$ \textbf{to} $d$}{
      Precompute vector $Q^{inner}_k$ \;
      Find such $x_k$, that $\left( Q^{inner}_k, \int_{-inf}^{x_k} \vec{f}(x_k) dx_k \right) = u_k$ \;
      Update $Q^{left}_{k}$ \;
    }

    \Return{$\vec{x}$}
    \caption{Algorithm to retrieve an exact sample $\vec{x}$ from the density $q_{\vec{\theta}}(\vec{x})$ represented in TT-format.}
    \label{alg:sampling}
  \end{algorithm}

  \section{Learning via Riemannian optimization}
  \label{sec:optimization}

  \paragraph{Loss function.}
  \label{sec:theory:losses}
  In practice, we do not know the true probability function $p$ and have access only to the dataset $\set{X} = \{\vec{x}_i\}_{i = 1}^{n}$ of i.i.d. samples $\vec{x}_i$ from density $p$. Thus, we need to optimize some loss function in order to get an approximation $q_{\vec{\theta}}$ of $p$. In the previous section we showed that we can explicitly calculate the partition function for the proposed tensor train-based model. Similarly, it is possible to calculate $L_2$ norm of the function represented in the  tensor-train format in just a $\O\bigl(d (r^3 m + r^2 m^2)\bigr)$ time. That allows us  to use an interesting loss, unusual in the density estimation context, computed by the direct calculation of $L_2$ distance between target distribution $p$ and approximation $q_{\vec{\theta}}$:
  \begin{EQA} \label{eq:loss-l2}
  \mathcal{L}(p, q_{\vec{\theta}}) & = & \int \bigl(p(\vec{x}) - q_{\vec{\theta}}(\vec{x}) \bigr)^2 dx
  \\
  & = & \int q_{\vec{\theta}}(\vec{x})^2 dx - 2 \E_{\vec{x} \sim p(\vec{x})} q_{\vec{\theta}}(\vec{x}) + const.
  \end{EQA}
  As only samples are available from density $p$, the expectation in~\eqref{eq:loss-l2} can be approximate with the Monte-Carlo estimate based on the samples from the dataset. If the expressive power of $q_{\vec{\theta}}$ is large enough, then minimum will be achieved near $p$ and thus will produce a good approximation to the true density function.

  \paragraph{Computation of loss function and its derivatives.}
  The second term in~\eqref{eq:loss-l2} is just an evaluation of the function at a given point. It was discussed in details in Section~\ref{sec:basis} and can be calculated in $\O(b d r^2 m)$ time, where $b$ is a batch size used in the stochastic optimization method. The first term in~\eqref{eq:loss-l2} is a quadratic function w.r.t. tensor $\alpha_{\vec{\theta}}$ and thus can be expressed in form $\langle \alpha_{\vec{\theta}}, D \alpha_{\vec{\theta}} \rangle$ with an appropriate choice of the linear operator $D$:
  \begin{EQA}[rcl]
    \int \bigl\langle \alpha_{\vec{\theta}}, \Phi(\vec{x}) \bigr\rangle^2 d \vec{x}
    & = & \int \Bigl\langle \alpha_{\vec{\theta}}, \bigl(\Phi(\vec{x}) \otimes \Phi(\vec{x}) \bigr) \alpha_{\vec{\theta}} \Bigr\rangle d \vec{x} \\
    & = & \left\langle \alpha_{\vec{\theta}}, \left( \int \Phi(\vec{x}) \otimes \Phi(\vec{x}) d \vec{x} \right) \alpha_{\vec{\theta}} \right\rangle,
  \end{EQA}
  where
  \begin{EQA}[l]
    \left[ \int \Phi(\vec{x}) \otimes \Phi(\vec{x}) d \vec{x} \right]_{i_1, j_1, \dots, i_d, j_d} \\
    = \int \Phi(\vec{x})_{i_1, \dots, i_d} \otimes \Phi(\vec{x})_{j_1, \dots, j_d} d \vec{x} \\
    = \int f_{i_1}(x_1) f_{j_1}(x_1) d x_1 \cdots \int f_{i_d}(x_d) f_{j_d}(x_d) d x_d \\
    = D_{i_1, j_1} \cdots D_{i_d, j_d}
  \end{EQA}
  and
  \begin{EQA}[l]
    D_{i, j} = \int f_i(x) f_j(x) dx.
  \end{EQA}
  Thus, tensor $\bigl[ \int \Phi(\vec{x}) \otimes \Phi(\vec{x}) d \vec{x} \bigr]$ is  rank-1 tensor being an outer product of $d$ matrices $D_{i, j}$. Thus, an application of it to the tensor $\alpha_{\theta}$ boils down to the multiplication of each core of $\alpha_{\theta}$ along the middle axis with the matrix $D$ followed by the inner product of two rank-$r$ tensors. It results in total complexity $\O\bigl(d m r^2 (m + r)\bigr)$. Let us note that an additional multiplication by $(m + r)$ does not significantly increase the computational cost of the optimization process as this term does not depend on the batch size $b$.

  \paragraph{Riemannian optimization and optimal step.}
  \label{sec:theory:optimization}
  In principle, the standard stochastic gradient descent methods or their variations can be used to train the proposed model. However, the representation of the model in tensor-train format and the fact, that the given loss function~\eqref{eq:loss-l2} is quadratic with respect to the tensor $\alpha$ allow for usage of more productive optimization methods. In this work, we suggest using Riemannian optimization, which is a promising tool for learning tensor-based models~\citep{Rakhuba2019,Steinlechner2016}.

  Riemannian optimization is a procedure to minimize some function $g$ defined on $\set{X}$ over some smooth manifold $\set{M} \subset \set{X}$:
  \begin{EQA}[l]
    \min_{\vec{X} \in \set{M}} g(\vec{X}).
  \end{EQA}
  The usual Riemannian optimization workflow consists of several steps:
  \begin{enumerate}
    \item Construction of \emph{tangential plane $\set{T}_{\vec{x}}(\set{M})$} to manifold $\set{M}$ at point $\vec{x}$. For tensor-train format, it has an efficiently computable expression, see~\citep{Rakhuba2019}.

    \item Projection $\mathbb{P}_{\set{T}_{\vec{X}}(\set{M})} \nabla g(\vec{X})$ of the true gradient $\nabla g(\vec{X})$ onto the tangent plane $\set{T}_{\vec{X}}(\set{M})$, which can be done efficiently using automatic differentiation.

    \item Gradient step in the tangent plane: $\vec{X}^{next} = \vec{X} + \alpha \mathbb{P}_{\set{T}_{\vec{X}}(\set{M})} \nabla g(\vec{X})$, where $\alpha$ is a learning rate.

    \item Retraction of a point $\vec{X}^{next}$ back onto $\set{M}$, which again can be efficiently approximated for tensor-train format.
  \end{enumerate}
  %
  By the construction, the tangent plane $\set{T}_{\vec{x}}(\set{M})$ is a linear space, and due to the fact that our loss function~\eqref{eq:loss-l2} is a quadratic function, we can find optimal $\alpha$ on each step as a minimal point of a parabola. Note that it is not true for the classical gradient descent in the space of TT-cores, as there will be a complex high degree polynomial dependence.

  More details on how to construct $\set{T}_{\vec{X}}(\set{M})$, how to represent it in the tensor-train format of doubled rank $2r$ and how to project the true gradient onto tangent space can be found in Supplementary Material~\ref{supp-subsec:riemannian} and in~\citep{Rakhuba2019}.

  \paragraph{Initialization.}
  \label{sec:theory:initialization}
  It is important to have good initialization for the gradient optimization methods in general as well as for the proposed Riemannian optimization approach.
  Rather efficient but straightforward initialization can be performed under the assumption of coordinate independence. Consider the case where
  \begin{EQA}[l]
    p(\vec{x}) = \prod_{i=1}^{d} p_i(x_i)
  \end{EQA}
  for some set of one-dimensional probability density functions $\{p_i\}_{i = 1}^d$. We can solve approximation problems
  \begin{EQA}[l]
    \label{eq:init-1d-task}
    p_i(x_i) \approx \bigl\langle \alpha_i, f(x_i) \bigr\rangle, \quad i = 1, \dots, d
  \end{EQA}
  independently in such a case and then consider $\alpha_{\vec{\theta}} = \alpha_1 \otimes \cdots \otimes \alpha_d$ as a rank-1 tensor-train initialization. Each approximation~\eqref{eq:init-1d-task} can be computed as the solution of a simple linear regression problem for the loss~\eqref{eq:loss-l2}.

  \section{Related work}
  \label{sec:related}

  One of the most famous non-parametric density estimation algorithms is the histogram approach~\citep{SCOTT1979}, which works only in very low dimensional settings (1-2 dimensions). Another famous method is the celebrated kernel density estimation approach~\citep{Scott1977} that is again known to perform poorly in high dimensions. 
  An important quality of these classical methods for approximating the distribution of low-dimensional data is their simplicity and intuitive behavior. However, many modern methods discussed below greatly improve the quality of density estimation which comes at the cost of much more sophisticated estimation procedures.

  The recent development of artificial neural networks gave birth to several new families of non-parametric density estimation. Generative-adversarial networks (GANs, \cite{goodfellow2014generative}) are methods of building the neural network capable of generating synthetic data close to the observed data. Although astonishing performance in real-life problems and the ability to learn and generalize extremely high-dimensional and complex data (images, videos and sound), this methods do not produce tractable (or even intractable) density functions and thus the applicability of these methods in statistical context is limited. Another similar approach is variational autoencoder (VAE, \cite{Kingma2013}). Unlike GANs, VAEs minimize the variational lower bound of the likelihood and thus can be used to approximate the unnormalized density functions, although the partition function is still intractable.

  There is also a great variety of methods, based on neural networks, that directly learn the density function: energy-based models~\citep{LeCun2006}, autoregressive density estimators~\cite{Ryder2018} and normalizing flows~\cite{Kobyzev2020}, among some others. Energy-based models learn the unnormalized density functions by maximizing the log-likelihood of the data and approximate the partition function by MCMC sampling. Thus, only approximate sampling from these models is available (via MCMC). Autoregressive models factor the probability distribution $p(\vec{x}) = \prod_{k=1}^{d} p(x_k \mid x_{s < k})$, parameterizing each factor with a neural network. Methods based on normalizing flows build smooth bijection of the target space with the latent space of the same size: $z = f_{\vec{\theta}}(x) \Rightarrow p(x) = p(z) \left| \frac{\partial f}{\partial x} \right|$. By setting the simple distribution of the latent variables (usually standard Gaussian) and assuming that the log-determinant of the function $f_{\vec{\theta}}$ can be efficiently calculated (guaranteed by choosing the appropriate neural network structure), the likelihood of the observed data can be directly maximized. This method was successfully applied to such complex tasks as face and speech generation~\citep{Kingma2018,Kim2020}. The recent development of NF models allows the use of more and more complex and less constrained models (see, for example, FFJORD~\citep{Grathwohl2018}), but at the same time, some of them lose the ability to sample from the trained model (BNAF~\citep{DeCao2019}).

  Another family of models that offers tractable query class is Probabilistic Circuits~\citep{choiprobabilistic} -- acyclic directional computational graphs that represent complex distributions as mixtures (sum nodes in the graph) and factorizations (product nodes in the graph) with simple tractable distributions, usually one-dimensional, in the leaves of this graphs (input nodes).
  To some extent, TTDE can be seen as a Probabilistic Circuit with basis functions being input nodes, and the graph structure implicitly defined in agreed with tensor-train format~\eqref{eq:tt-of-alpha}, which would give us, although predefined, \emph{exponentially large} model structure.
  Also, TTDE admits negative weights, which potentially beneficially affects expressiveness~\citep{dennis2016algorithms}.

  After the initial publication, we became aware of other methods that are also based on low-rank tensor decompositions~\cite{KargasSF18,KargasS19,amiridi2021lowrank}. A very close line of research uses Canonical Polyadic Decomposition to approximate probability tensor of random vectors over finite alphabet~\cite{amiridi2021lowrank}. In the work~\cite{KargasS19} the same decomposition is applied to the Fourier expansion coefficients to approximate the continuous distribution. Although in these works theoretical guarantees of identifiability of distributions (for distributions of low enough rank) have been obtained, in practice, tensor-train decomposition tends to be more stable and exponentially more expressive~\cite{Khrulkov2017} with the same rank, thus detailed comparison of different tensor decompositions in this scenario is an interesting topic for future research.

  \section{Experiments}
  \label{sec:experiments}
  We evaluate the performance of the presented TTDE method on several model and real-world datasets. All the code to reproduce the results of experiments can be found via \url{https://github.com/stat-ml/TTDE}. In all the experiments basis function set consists of B-splines of degree 2 with knots uniformly distributed over the support of the considered distributions. The support is known precisely for the simulated examples as we know exactly the target distribution, and corresponding lower and upper bounds are extracted from all given samples for the unknown distributions of the real-world datasets.

  \subsection{Toy examples}
\begin{figure}[t]
    \centering
    \includegraphics[width=.7\linewidth]{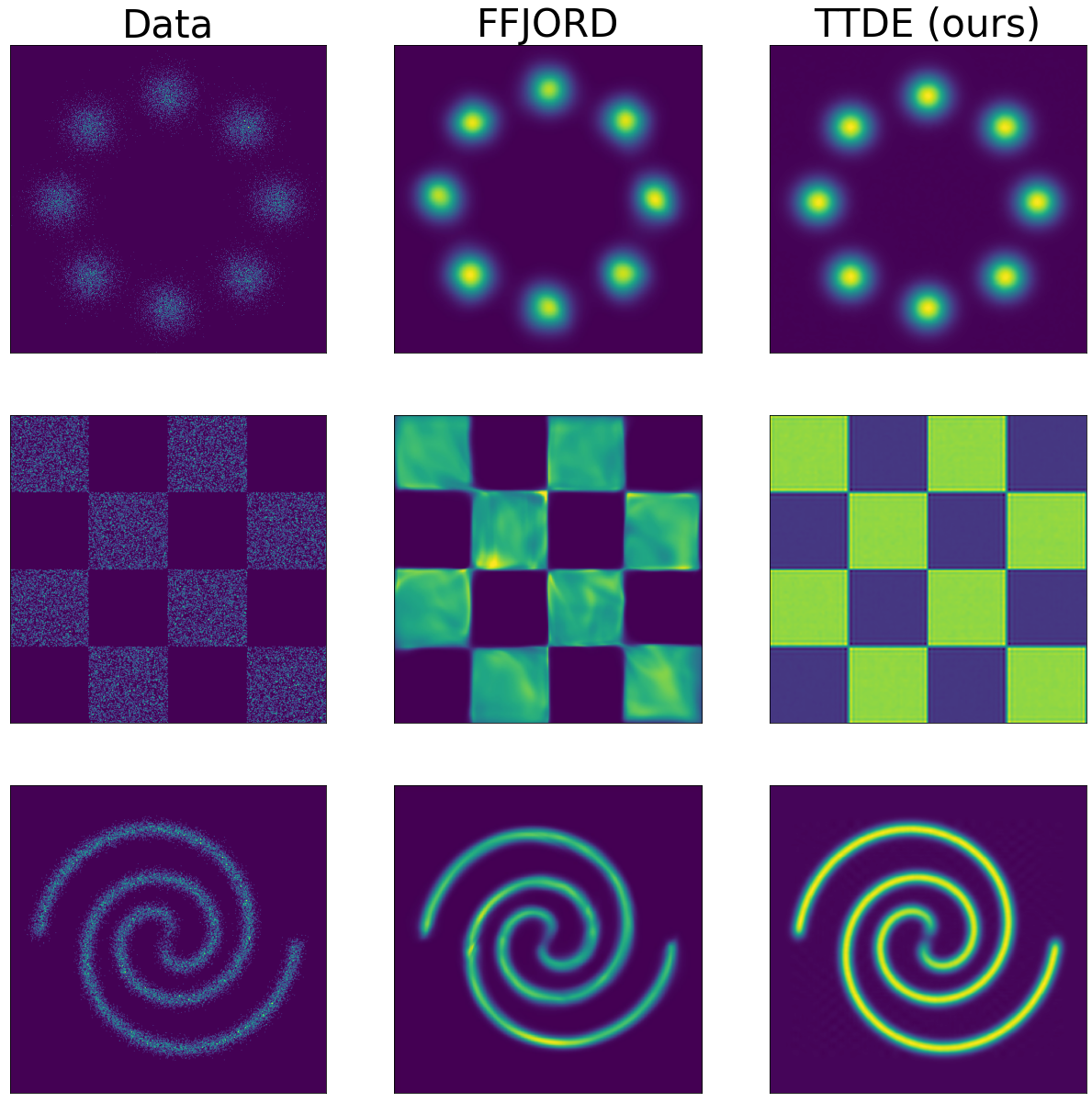}
    \caption{Comparison of TTDE and FFJORD models on 2-dimensional toy distributions.} \label{fig:toy}
  \end{figure}

  We start the presentation of our results with the several classical $2$-dimensional examples, which are presented on the Figure~\ref{fig:toy}. We clearly observe that TTDE doesn't have any problems with complex shape of the distribution, its multimodality or discontinuity. Moreover, the visual comparison shows clear superiority of TTDE over state-of-the-art normalizing flow approach FFJORD~\citep{Grathwohl2018}.

  \subsection{Model datasets and hyperparameter selection}
\begin{figure*}[t]
    \centering
    \includegraphics[width=.65\linewidth]{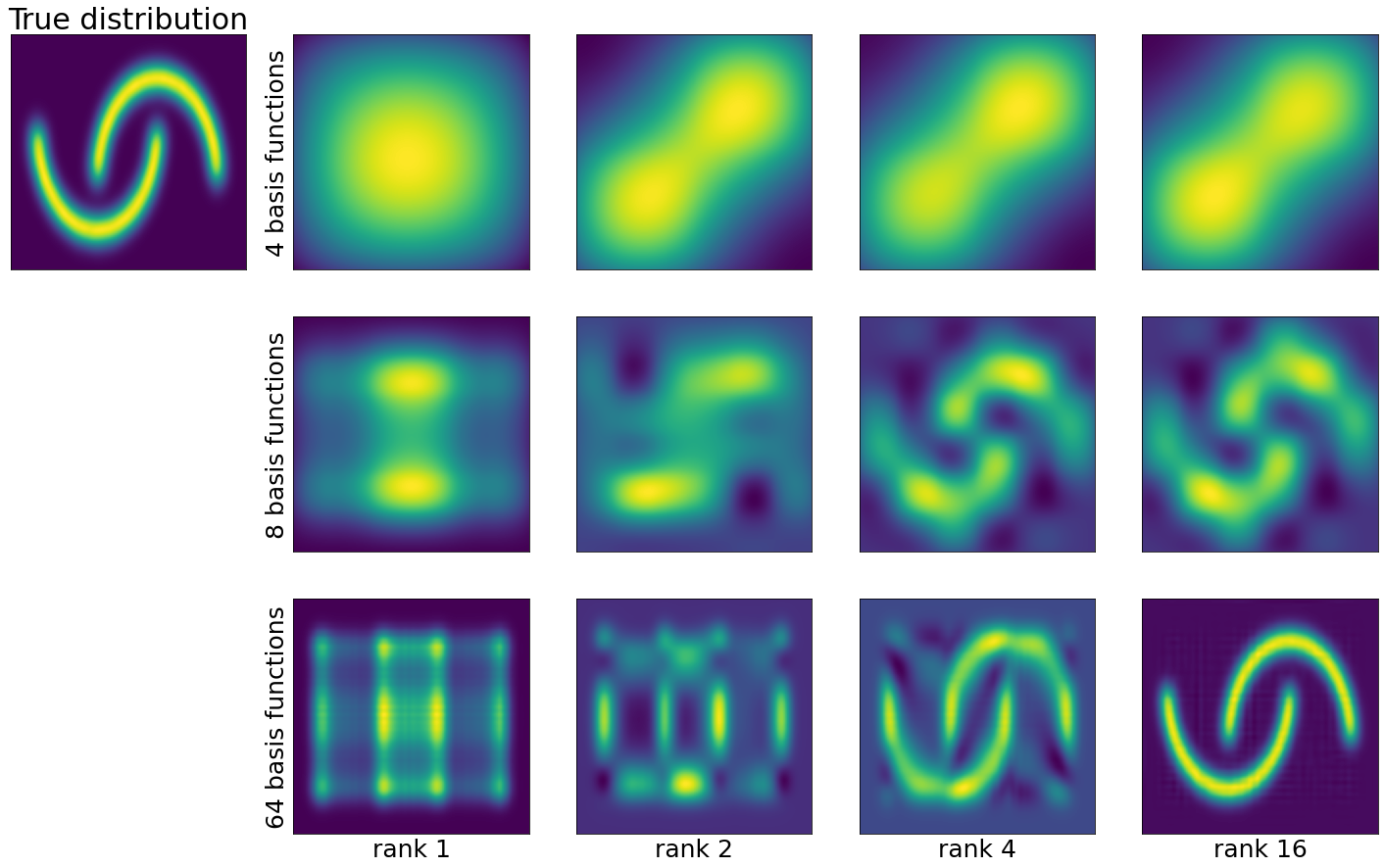}
    \caption{Approximations of ``two moons'' distribution by TTDE for different basis function set sizes and TT-ranks.} \label{fig:rank-basis-moons}
  \end{figure*}

  An important feature of our model is great interpretability of the model hyperparameters: basis size and tensor-train rank. The basis size corresponds to the resolution of the approximation. It acts similarly to the number of bins in the very large multidimensional histogram. Rank of the tensor-train decomposition corresponds to the expressive power of the model, i.e. how complex distributions can be built for the given basis size. The dependence of the trained density on both hyperparameters is shown on Figure~\ref{fig:rank-basis-moons} for the celebrated two moons dataset. We clearly observe the behavior discussed above. Interestingly, if the rank of tensor-train decomposition is not large enough, the method tries to somehow cope with it, adding symmetric artifacts to the distribution.

\begin{figure}[t]
    \centering
    \includegraphics[width=.35\textwidth]{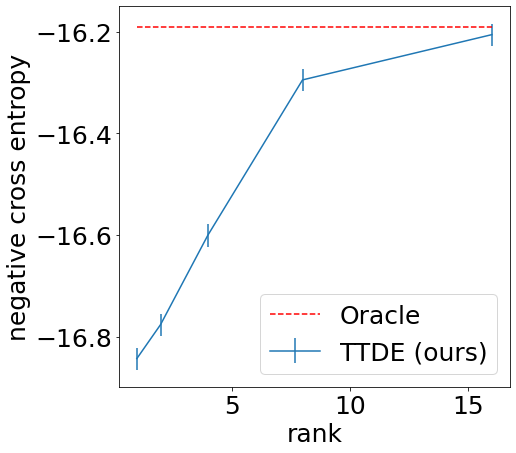}
    \caption{Dependence of the approximation quality of the mixture of 128 Gaussians in 8 dimensional space on the rank of the TT decomposition.} \label{fig:rank-asymmetric-8d}
  \end{figure}

  We additionally explore TTDE properties dependence on the rank of the model for the 8-dimensional mixture of 128 Gaussians located in the random corners of 8-dimensional unit cube. We present the dependence of the cross-entropy on the rank of the tensor-train decomposition on Figure~\ref{fig:rank-asymmetric-8d}. We observe an expected behavior: the higher is the rank, the higher is the cross entropy with the true distribution. Importantly, already the approximation of the rank 16 is enough to almost perfectly match this very complex distribution.

  \begin{table}[t]
    \centering
    \begin{tabular}{|l|l|l|}
      \hline
      & Random init. & Rank-1 init. \\
      \hline
      Adam       & 5                   & 11                     \\
      \hline
      Riemannian & 12                  & 32                     \\
      \hline
    \end{tabular}
    \caption{Experiment with mixture of 7 Gaussians in 3D with additional dimensions containing only noise. We report the maximum dimensionality for which approximation of the density converges to the true one for different initialization settings and optimization methods used.}
    \label{tab:max-dimension}
  \end{table}

  \subsection{Importance of initialization.}
  \label{subsec:importance}
  To show the importance of the proper initialization and optimization method, we tested our method on the mixture of seven identical $n$-dimensional Gaussian mixtures with identity covariance matrix and located in the 7 corners (all except one) of the 3-dimensional cube in the first 3 dimensions of the space (see the illustration in Supplementary Material). This is rather simple distribution having rank 2. We trained four different models: with rank-1 initialization (see Section~\ref{sec:theory:initialization}) or with random initialization, and with proposed Riemannian optimizer (see Section~\ref{sec:theory:optimization}) or with the standard Adam optimizer. We set the rank of the approximation equal to 4 (slightly larger, than the true rank of the target distribution).

  For each combination of initialization and optimization method we report the largest dimensionality of the data for which the corresponding method successfully converges to correct solution. The resulting numbers are shown in Table~\ref{tab:max-dimension}. We see that for random initialization Riemannian optimization allows to achieve correct results in much higher dimension than Adam (12 vs 5). The usage of the proposed 1d-initialization procedure allows to significantly boost the results for both optimizers. The clear winner is properly initialized model with Riemannian optimization which is capable to learn $32-$ dimensional distribution. Thus if we want to apply this algorithm in high dimensional settings, the proposed initialization and the right choice of optimization technique are vital.

  \subsection{Real-world data}
  \label{subsec:real-world}

  \begin{table*}[h]
    \centering
    \caption{Average log-likelihood for several neural-network-based models on tabular UCI datasets. Gaussian fitted to the training data is reported as a baseline. \\
    *On Hepmass and Miniboone datasets, which has the lowest number of training examples (300k and 30k respectively), we observe heavy overfitting. Lack of regularizations for the new model leads to poor results. Thus, it is an important direction for further development of the TTDE. }
    \label{tab:sqrttde-results}
    \begin{tabular}{cccccc}
      \toprule 
      & POWER & GAS & HEPMASS & MINIBOONE & BSDS300 \\
      \midrule 
      Dataset dimensionality & 6 &  8 & 21 & 43 & 64 \\
      \midrule
      Gaussians & -7.74 & -3.58 & -27.93 & -37.24 &  96.67 \\
      MADE      & -3.08 &  3.56 & -20.98 & -15.59 & 148.85 \\
      Real NVP  &  0.17 &  8.33 & -18.71 & -13.84 & 153.28 \\
      Glow      &  0.17 &  8.15 & -18.92 & -11.35 & 155.07 \\
      FFJORD    &  0.46 &  8.59 & \bfseries -14.92 & \bfseries -10.43 & \bfseries 157.40 \\
      \midrule
      Squared TTDE (ours) & \bfseries 0.46 & \bfseries 8.93 & $-21.34^*$ & $-28.77^*$ & 143.30 \\
      \bottomrule 
    \end{tabular}
  \end{table*}

  \paragraph{Methods and data.}
  To show the applicability of our method to the real-world tasks, we compare computational performance and quality of approximation of TTDE and squared-TTDE with several methods from the Normalizing Flows family as they have similar capabilities (see Table~\ref{tab:comp}), namely Glow~\citep{Kingma2018}, RealNVP~\citep{Dinh2016}, MAF~\citep{Papamakarios2017} and FFJORD~\citep{Grathwohl2018}. We perform comparison on five tabular datasets from UCI dataset collection preprocessed as in~\cite{papamakarios2017masked}.

  \paragraph{Quality measure.}
  Due to the fact that the function represented in the tensor-train format does not have to be a valid probability density function i.e. potentially there could be areas of small negative values due to approximation error, direct comparison in terms of model likelihood is not available. On the other hand, comparison in terms of the loss~\eqref{eq:loss-l2} is not a good choice as well. Firstly, it is problematic to calculate it for the normalizing flow models in $6$-dimensional space. Secondly, it is not fair to compare the model that directly optimizes this loss (our model) with the models, that are trained using completely different discrepancy measures. Because of that, we decided to compare models based on the quality of generated samples. Namely, we decided to measure the \emph{sliced total variation} between the samples from the validation set and the samples acquired from the model.

  Total variation is a classical measure of discrepancy between two distributions
  \begin{EQA}[l]
    TV(p ~\|~ q) = \int \bigl|p(x) - q(x)\bigr| dx.
  \end{EQA}
  In spite of the simplicity of its formulation, the computation of the integral above in high dimension is hard and potentially non-accurate. That's why we introduce the sliced version of the total variation by employing a simple idea how to apply it to the multidimensional case when the true distribution $p(x)$ is not available. Instead of integrating over the whole space, we can average total variations for many random $1$-dimensional projections:
  \begin{EQA}[l]
    STV(p ~\|~ q) = \E_{P} \int \bigl|P[p](x) - P[q](x)\bigr| dx,
  \end{EQA}
  where $P$ is a random projector on one dimensional space.
  In our experiments we firstly generate samples from all testing models. Then several times we generate projection of validation set and all generated sets on a random $1$-dimensional plane. Then we calculate $1$-dimensional approximations of distributions using kernel density estimation (which almost exactly replicates true underlying distribution due to massive amounts of generated points) and then we calculated TV of this $1$-dimensional functions. This procedure is summarized in Algorithm~\ref{alg:STV}.

  \begin{algorithm}[h]
    \SetAlgoLined
    \KwResult{Sliced total variation for two sample sets $X^{(1)} =\{x^{(1)}_i \in \R^d\}_{i=1}^{N}$ and $X^{(2)} =\{x^{(2)}_i \in \R^d\}_{i=1}^{N}$}
    Choose random $1-dimensional$ hyperplane $l$ \;
    Project $X^{(1)}$ and $X^{(2)}$ onto $l$ \;
    Build approximations $p_1$ and $p_2$ from projected samples\;
    Approximate $STV = \int_{\R} |p_1(y) - p_2(y)| dy$ using any $1$-dimensional numerical integration method \;
    \Return{$STV$}\;
    \caption{Calculation of sliced total variation between two distributions based on samples from them.}
    \label{alg:STV}
  \end{algorithm}

\begin{figure}[t]
    \centering
    \includegraphics[width=.7\linewidth]{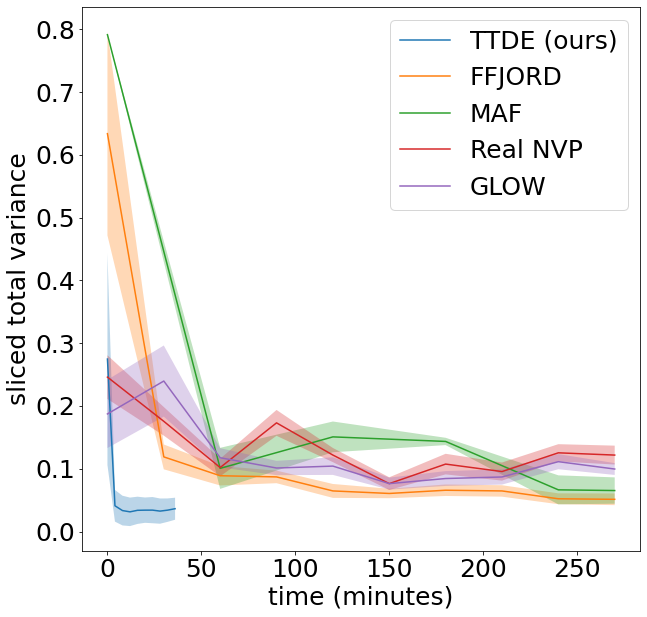}
    \caption{Dependence of the sliced total variation w.r.t. the training time for models  trained on  $6$-dimensional UCI POWER dataset.} \label{fig:ffjord-tt-tv}
  \end{figure}

\begin{figure}[t]
    \centering
    \includegraphics[width=.7\linewidth]{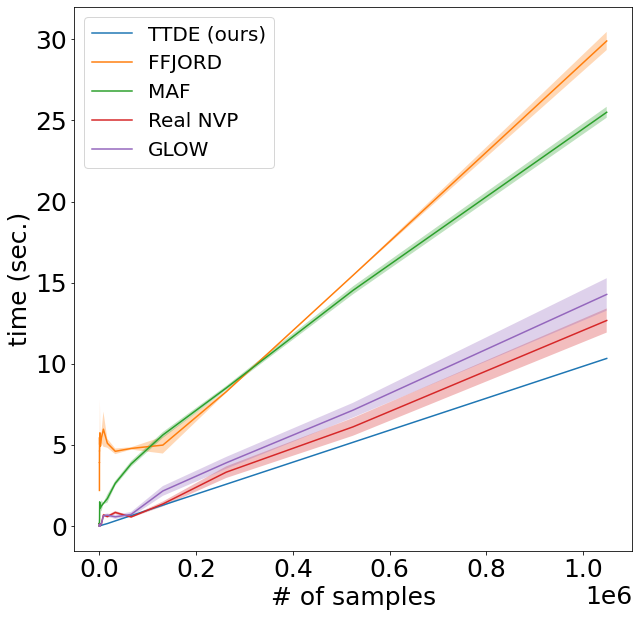}
    \caption{Dependence of the sampling time w.r.t. the number of samples to be generated for $6$-dimensional space for models trained on UCI POWER dataset. Our model outperforms its competitors and shows $2.6$, $2.5$, $1.4$ and $1.2$ times speedups compared to FFJORD, MAF, GLOW and Real NVP respectively.} \label{fig:ffjord-tt-sampling}
  \end{figure}

  \paragraph{Results.}
  We show in our experiments that the proposed model significantly outperforms (by an order of magnitude) all presented neural-network based models in terms of the convergence rate to the optimal value, see Figure~\ref{fig:ffjord-tt-tv}. The same applies to the speed of sampling, see Figure~\ref{fig:ffjord-tt-sampling}, where for the batch size of $2^{20}$ we outperform two most powerful baseline models: FFJORD and MAF (speedup of $2.8$ and $2.5$ times respectively) and slightly outperform GLOW and Real NVP ($1.4$ and $1.2$ times speedup respectively).

  In table~\ref{tab:sqrttde-results} we report log-likelihoods achieved with our squared-TTDE model trained with NLL loss and compare them with several normalizing flow models. Our model manages to outperform presented competitors on two low-dimensional datasets (POWER and GAS). On the 64-dimensional BSDS300 dataset, our model performs worse than powerful network-based models, although by not much, while still providing tractable query class. On Hepmass and Miniboone datasets, which have the lowest number of training examples (300k and 30k respectively), we observe heavy overfitting. We think that development of regularization techniques for the TTDE will allow to obtain better results. All the details of experiments are specified in Supplementary Materials~\ref{supp-subsec:setups}.

  \section{Conclusion}
  \label{sec:conclusion}
  This work shows that approximation based on the tensor-train decomposition is a promising method of density estimation. It offers such a set of different features and possibilities (tractable partition function, exact sampling, exact marginals, and cumulative densities), which were not previously accessible all at the same time for other methods. This method's ability to work in medium dimensionality is very promising and paves the way to accurate density estimation in high dimensions.

  \acknowledgements
  The research was supported by the Russian Science Foundation grant 20-71-10135.

  \bibliography{ms.bib}


\end{document}


\maketitle

\appendix

\section{Supplementary Material}
\label{sec:suppl}

\subsection{Details of the Experiments}
\label{subsec:setups}
  In all the experiments set of B-splines of degree $2$ with knots uniformly distributed over the distribution support was used as basis functions for rank-$1$ feature maps. In all the experiments, Riemannian optimization with the optimal learning rate was used if not stated otherwise. In Section~\ref{main-subsec:importance} Adam optimizer from PyTorch was used with default parameters. Sampling from TTDE was performed with 30 binary search iterations. In all the experiments, the batch size was $2^{10}$ elements per iteration. For all toy and model examples, we used infinite data generators. In real-world data experiments in Section~\ref{main-subsec:real-world}, the rank $r$ of the TTDE was $64$, and the number of basis functions $m$ was $128$. Implementation of FFJORD was taken from \url{https://github.com/rtqichen/ffjord}, implementations of GLOW, Real NVP and MAF were taken from \url{https://github.com/ikostrikov/pytorch-flows} and used with parameters recommended by authors.
  
\begin{figure}[t]
    \centering
    \includegraphics[width=.9\linewidth]{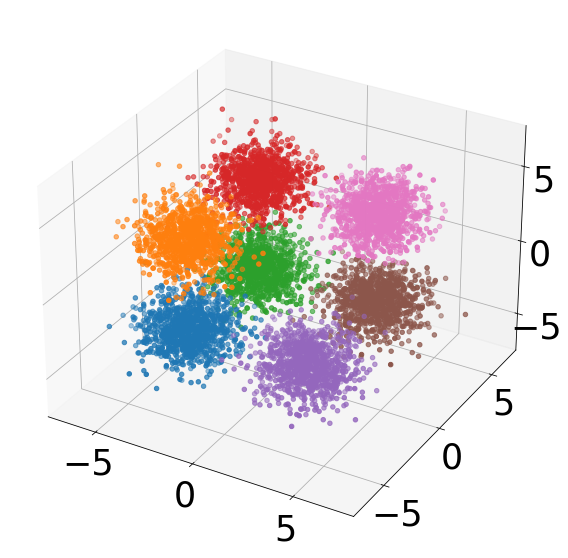}
    \caption{Visualization of the first three dimensions of the model distribution used in this work. It consists of 7 identical Gaussian distributions shown with different colours. Other $d - 3$ dimensions are standard Gaussian noise. }
  \label{fig:distrs:asym}
  \end{figure}

  The first three components of the distribution used in Section~\ref{main-subsec:importance} are depicted on Figure~\ref{fig:distrs:asym}. Other $d - 3$ components are standard Gaussian noise.
  
\subsection{Existing measures of discrepancy}
\label{sup:losses}
  
  $\mathcal{KL}$-divergence is a popular measure of discrepancy between two distributions, and, during training, is presented in the form of the maximum likelihood problem:
  \begin{EQA}[l]
    \mathcal{KL}(p ~\|~ q_{\vec{\theta}}) = -\E_{\vec{x} \sim p(\vec{x})} \log q_{\vec{\theta}}(\vec{x}) + const.
  \end{EQA}
  Different models were built to optimize this kind of discrepancy (including autoregressive models~\citep{Ryder2018}, normalizing flows~\citep{Kobyzev2020}, energy-based models~\citep{LeCun2006}). The main downside of the maximization of the likelihood is that it explicitly depends on the partition function of the approximation. Thus either the models should be constructed in such a way that the partition function could be efficiently calculated, or expensive Monte-Carlo methods should be used to approximate it during the optimization. We can not use $\mathcal{KL}$-divergence to train TTDE because, although it has a tractable partition function, function in tensor-train format is not guaranteed to be positive.

  Fisher discrepancy loss does not depend on the partition function of the approximation:
  \begin{EQA}[ll]
    & \mathcal{L}(p, q_{\vec{\theta}}) \\
    = & \E_{\vec{x} \sim p(\vec{x})} \norm{\nabla \log p(\vec{x}) - \nabla \log q_{\vec{\theta}}(\vec{x})}^2 \\
    = & \E_{\vec{x} \sim p(\vec{x})} \norm{\nabla \log q_{\vec{\theta}}(\vec{x})}^2 - \E_{\vec{x} \sim p(\vec{x})} \Delta \log q_{\vec{\theta}}(\vec{x}) + const.
  \end{EQA}
  Here $const$ depends only on $p$ and does not depends on $q_{\vec{\theta}}$. Because of the gradient of the logarithm, the normalization constant cancels out. The downside of this loss is that for complex models like neural networks, the Laplace operator is hard to calculate from both computational and numerical stability points. This loss can be used to train TTDE if we parameterize $\log p(x)$ instead of parameterizing $p(x)$ with the tensor-train model. However, in that case, we would lose the ability to calculate partition function and cumulative density function and thus would not be able to exact sample.

  Different versions of adversarial loss were created and successfully used to learn complex distributions like images or speech. They use a separate model as a critic during the training process. Consider the following two variants for vanilla GAN and WGAN, respectively:
  \begin{EQA}
    && \mathcal{L}(p, q_\theta) =\\
     & = & \max_D \left\{ \E_{\vec{x} \sim p(\vec{x})} \left[ \log D(\vec{x}) \right] + \E_{\vec{z} \sim p_{\vec{z}}(\vec{z})} \left[ \log(1 - D(G(\vec{z}))) \right] \right\},
    \\
    && \mathcal{L}(p, q_\theta) =\\
    & = & \max_D \left\{ \E_{\vec{x} \sim p(\vec{x})} \left[ D(\vec{x}) \right] - \E_{\vec{z} \sim p_{\vec{z}}(\vec{z})} \left[ D(G(\vec{z})) \right] \right\}.
  \end{EQA}
  Generator $G$ maps latent variable $\vec{z}$ with known distribution $p_{\vec{z}}$ to the sample space $\vec{x}$, and discriminator $D$ tries to distinguish between real samples and generated samples. In these cases, $q_{\vec{\theta}}$ is defined implicitly. Separate choice of the critic architecture, instability of the optimization of the min-max problem, and the intractability of the implicit density function $q_{\vec{\theta}}$ are the problems that come with the power of adversarial models.

\subsection{Riemannian Optimization}
\label{subsec:riemannian}
  \paragraph{Orthogonalization.} 
  Left- and right-orthogonalization of the tensor-train decomposition is three sets of matrices: set of left-orthogonal cores $\vec{U}_1, \dots, \vec{U}_d$, set of right-orthogonal cores $\vec{V}_1, \dots, \vec{V}_d$ and set of unrestricted cores $\vec{S}_1, \dots, \vec{S}_d$, such that
  \begin{EQA}[l]
  \label{eq:orthogonalization}
    G_1 \times^1_2 \cdots \times^1_{d} G_d = \\ 
    \vec{U}_1 \times^1_2 \cdots \vec{U}_{i-1} \times^1_{i} \vec{S}_i \times^1_{i+1} \vec{V}_{i + 2} \cdots \times^1_{d} \vec{V}_{d}
  \end{EQA}
  for each $i$, where left-orthogonality means
  \begin{EQA}[l]
    \langle U_k[:, :, i], U_k[:, :, j] \rangle = \delta^i_j,
  \end{EQA}
  and right-orthogonality means
  \begin{EQA}[l]
    \langle V_k[i, :, :], V_k[j, :, :] \rangle = \delta^i_j.
  \end{EQA}  
  
  \paragraph{Tangent space.}
  Given the left- and right-orthogonalization of the given tensor-train decomposition of tensor $X$, tangent space $\set{T}_{\vec{X}}(\set{M})$ in that point could be constructed as follows:
  \begin{EQA}[l]
    \set{T}_{\vec{X}}(\set{M}) = \\
    \left\{ 
      \begin{array}{c}
         T = U_1 \times^1_2 \cdots U_{i - 1} \times^1_i S^{\delta}_i \times^1_{i+1} V_{i + 1} \cdots \times^1_d V_d, \\
        \text{where } 1 \leq i \leq d,\; S^{\delta}_i \in \R^{r_{i - 1} \times m \times r_i} 
      \end{array} 
    \right\}.
  \end{EQA}
  Although tensor $T$ is presented as a sum of $d$ tensors of rank $r$, they share common cores. Because of that, $T$ can be represented with rank $2r$:
  \begin{EQA}[l]
    T = \\
      \left[ \begin{array}{cc} U_1 & S^{\delta}_1 \end{array} \right]
      \left[ \begin{array}{cc} V_2 & \\ S^\delta_2 & U_2 \end{array} \right] \cdots 
      \left[ \begin{array}{cc} V_{d-1} & \\ S^\delta_{d-1} & U_{d-1} \end{array} \right] 
      \left[ \begin{array}{c} S^{\delta}_d \\ V_d\end{array} \right]
  \end{EQA}
  
  \paragraph{Automatic differentiation.}
  If we define operator
  \begin{EQA}[l]
    T_{\vec{X}}(S^{\delta}_1, \cdots, S^{\delta}_d) = \sum_{i=1}^d U_1 \times^1_2 \cdots U_{i - 1} \times^1_i S^{\delta}_i \times^1_{i+1} V_{i + 1}
  \end{EQA}
  that maps delta-cores $\left\{ S^\delta_i \right \}_{i=1}^{d}$ into the point in tangent plane, then for any function $g(\vec{X}) : \R^{n_1 \times \cdots \times n_d} \rightarrow \R$ projection of the true gradient $\nabla(\vec{X})$ onto the tangent plane $\set{T}_{\vec{X}}(\set{M})$ could be efficiently calculated as follows:
  \begin{EQA}[l]
    \mathbb{P}_{\set{T}_{\vec{X}}(\set{M})} \nabla g(\vec{X}) = T_{\vec{X}}(\tilde{S}^\delta_1, \cdots, \tilde{S}^\delta_d),
  \end{EQA}
  where 
  \begin{EQA}[l]
    \tilde{S}^\delta_i = \left. \frac{\partial}{\partial S^\delta_i} g(T_{\vec{X}}(S^\delta_1, \cdots, S^\delta_d) \right|_{S^\delta_1 = S_1, S^\delta_2 = \vec{O}_2, \cdots, S^\delta_d = \vec{O}_d}.
  \end{EQA}
  Here $S_1$ is defined in~\eqref{eq:orthogonalization}, and $\vec{O}_i$ is core with all elements equal to zero. All $\tilde{S}^\delta_i$ could be calculated using automatic differentiation of function $g \circ T_{\vec{X}}$.

\bibliography{supplement.bib}

\makeatletter\@input{xx.tex}\makeatother